\def\year{2020}\relax
\documentclass[letterpaper]{article} 
\usepackage{aaai20}  
\usepackage{times}  
\usepackage{helvet} 
\usepackage{courier}  
\usepackage[hyphens]{url}  
\usepackage{graphicx} 
\usepackage{graphicx}  
\usepackage[utf8]{inputenc}
\usepackage{amsmath}
\usepackage{tikz}
\usepackage{url}
\usetikzlibrary{bayesnet}
\usetikzlibrary{positioning}
\newcommand{\figref}[1]{\figurename~\ref{fig:#1}} 

\title{A Critical Look at the Applicability of Markov Logic Networks for Music Signal Analysis}
\author{Johan Pauwels, György Fazekas, Mark~B. Sandler\\ 
Centre for Digital Music\\
Queen Mary University of London\\
j.pauwels@qmul.ac.uk 
}
 
\begin{document}

\maketitle

\begin{abstract}
In recent years, Markov logic networks (MLNs) have been proposed as a potentially useful paradigm for music signal analysis. Because all hidden Markov models can be reformulated as MLNs, the latter can provide an all-encompassing framework that reuses and extends previous work in the field. However, just because it is theoretically possible to reformulate previous work as MLNs, does not mean that it is advantageous. In this paper, we analyse some proposed examples of MLNs for musical analysis and consider their practical disadvantages when compared to formulating the same musical dependence relationships as (dynamic) Bayesian networks. We argue that a number of practical hurdles such as the lack of support for sequences and for arbitrary continuous probability distributions make MLNs less than ideal for the proposed musical applications, both in terms of easy of formulation and computational requirements due to their required inference algorithms. These conclusions are not specific to music, but apply to other fields as well, especially when sequential data with continuous observations is involved. 
Finally, we show that the ideas underlying the proposed examples can be expressed perfectly well in the more commonly used framework of (dynamic) Bayesian networks.
\end{abstract}

\section{Introduction}
Markov logic networks~\cite{richardson2006ml} form a part of the broader field of statistical relational learning~\cite{getoor2007itsrl}, which aims to combine complex relational information with probabilistic uncertainty. In the case of MLNs, the relations are expressed through first-order logic (FOL), and the uncertainty as weights assigned to those logic formulas. This way, hard logical rules in a standard FOL knowledge base can be relaxed and an order of importance imposed on the separate logical statements.

One particularly highly structured domain is music. Humans can label music according to a variety of aspects: tempo, harmony, mood, etc. All these labels are strongly related, and if they are time-varying, they are strongly correlated through time too. Trying to exploit those relationships for automatic music labelling is therefore a common research topic. So far, the majority of interdependency models have been constructed using hidden Markov models (HMMs) or other (dynamic) Bayesian networks ([D]BNs)~\cite{mauch2010taslp,pauwels2013ismir,pauwels2014jnmr}, but Markov logic networks have been proposed~\cite{papadopoulos2012ismir,papadopoulos2013icassp,papadopoulos2017taslp} as an alternative framework. 

One of the advantages of MLNs is that all HMMs can be reformulated as MLNs. Therefore previous work can easily be reused and be extended upon. However, just because it is theoretically possible to approach a problem as a MLN, does not mean that it is advantageous to do so. Papadopoulos and Tzanetakis illustrate the feasability of MLNs for music signal analysis by listing a number of example systems~\cite{papadopoulos2012ismir,papadopoulos2013icassp,papadopoulos2017taslp} that produce the chord label sequence corresponding to a music recording. Yet they do not specifically address why MLNs are preferable to HMMs or, more generally, Bayesian networks. On the other hand, the use of MLNs for those specific examples involves extra complications, which we believe to deserve a more thorough discussion.

In this paper, we try to address these issues, such that a better insight into the advantages and disadvantages of MLNs can be obtained. We look at alternative formulations of some MLN systems, and compare them on their ease of use, so our approach is purely theoretical because the numerical results are expected to be the same.

\section{Background}

\begin{figure*}[t]
\centering
\includegraphics[width=0.68\textwidth]{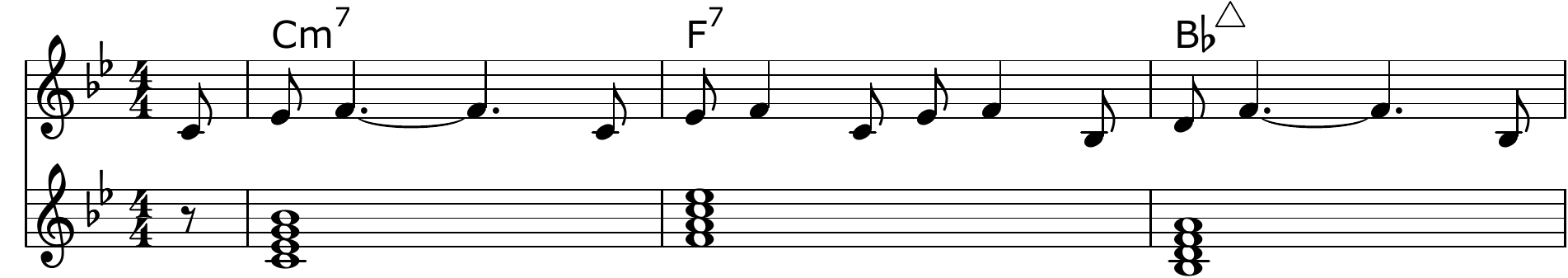} 
\includegraphics[width=0.68\textwidth]{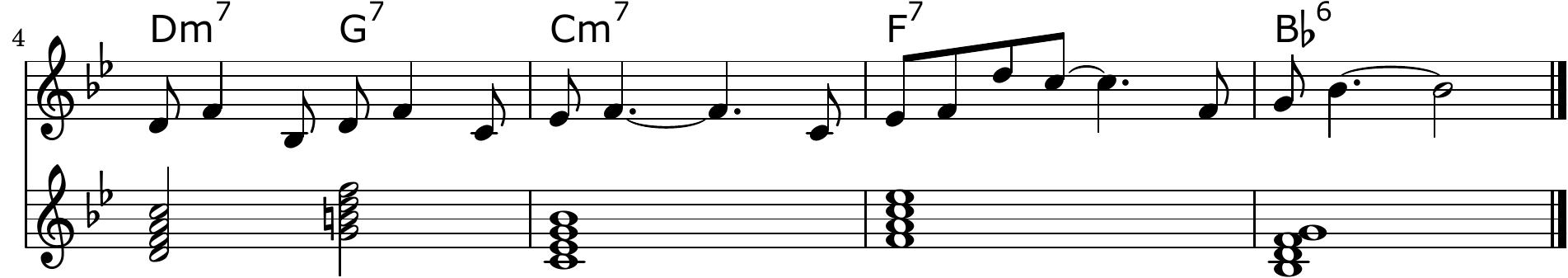} 
\caption{Sheet music example (``Perdido'' by Juan Tizol). Chord symbols can be seen above the staves and compactly represent and label the group of notes below them.\label{fig:sheet-music}}
\end{figure*}

Since Markov logic networks integrate the fields of graphical modelling and first-order logic, a great number of different concepts need to be defined. Furthermore, the application domain of music signal analysis comes with its own vocabulary, combining music theory and signal processing. Due to the limited space available here, and our goal of creating insight into their specific use-cases, we will describe the necessary concepts in an informal, intuitive way with an emphasis on their distinctions. For more thorough and technical definitions, the references should be consulted.

\subsection{Music Signal Analysis}

The field of music signal analysis aims to give machines the ability to recognise advanced musical concepts in audio recordings. These concepts can be expressed in natural language, such as the mood or genre of a music piece, or can be part of the symbolic language known as music notation. 
Examples of the latter are notes, chords and keys.

The usecase in this paper is the automatic recognition of chords from audio. The easiest way to explain chords is to start from sheet music, of which an example can be seen in \figref{sheet-music}. Sheet music can be seen a set of instructions for a musician on how to play a piece of music. 
Since a performance involves a fair amount of human interpretation, many different audio signals can be produced from the same sheet music.

The black stemmed dots in sheet music represent musical notes, displayed on five parallel lines (called a stave). The vertical axis represents pitch height, and the horizontal axis time, read from left to right. The relative duration of the notes is indicated by the shape of their stems. Staves connected by a vertical line at the beginning are played at the same time, and notes that appear in the same horizontal position (in any of the staves) are played concurrently. In our example, most of the time five notes are played together.

Chords are defined by \citeauthor{grovemusic} as ``the simultaneous sounding of two or more notes''. They are represented by the text-like representation above the staves. The complicated part about defining chords is that not all simultaneously played notes contribute to a chord. Some notes can be deemed to belong exclusively to the melody, or not significant enough to contribute to the chord. Deciding which notes get grouped together to form a chord is one of the challenges in chord recognition, and the reason why it is not trivial even when starting from sheet music. Of course, when the input is an audio signal, this complexity gets compounded.

In practice, chord recognition from audio is rarely based on recognising notes~\cite{pauwels2019ismir}. Instead, chords are directly obtained from a time-frequency representation that is processed with a sliding window. The resulting output is therefore not sheet music with chord symbols as in our example, but a list of text-encoded chord symbols with their associated timings in seconds. In order to obtain temporal consistency over the separately observed sliding windows, probabilistic graphical models are often used to smooth out the windowed output~\cite{pauwels2019ismir}. Statistical relational frameworks could provide an alternative formulation that has more potential to exploit the strong relationships between different musical concepts~\cite{papadopoulos2017taslp}.

\subsection{Markov Logic Networks}

Probabilistic graphical models are a way to graphically represent (in)dependence relationships between random variables. We can discern two main approaches: Bayesian networks (BN)~\cite[chapter 10]{murphy2012mlapp} and Markov networks (MN)~\cite[chapter 19]{murphy2012mlapp}. The latter are also known as Markov random fields. Both factorise the joint probability distribution of a set of random variables into a product of more specific distributions. For BNs, these factors are by definition conditional probabilities, whereas for MN the factors can be arbitrary non-negative functions. Because the factors in a BN are conditional probabilities, no normalisation is needed (a product of probabilities is always a probability). In MNs on the other hand, a normalisation term is needed such that all the factors sum to one. Because of this normalisation term over all factors, a MN is not localised and therefore it is harder to perform inference and learning. Its advantage, however, is its greater expressive power.

The difference between BNs and MNs also has consequences for their graphical representations. In both cases, variables are represented by nodes in a network. A BN is represented as a directed acyclic graph (DAG), because conditional probabilities are directional and non-circular. Conditional dependencies are represented by edges from the conditional variables to their dependent variables. A MN is represented as an undirected graph (UG), possibly with loops. Variables that appear in the same factor are densely interconnected by undirected edges. The collection of edges for such densely interconnected nodes is called a clique.

The mapping between factorisation and graph is not univocal though, one graph can represent multiple different factorisations. Furthermore, a graph is more conservative about stating independences than a factorisation: sometimes two variables are independent in the factorisation, even if the graphs says they are not. Never does a graph represent an independence that is not in the factorisation. The case when a factorisation and a graph represent exactly the same set of independences is called a perfect map.

The factors in a MN could be conditional probabilities as well, and in that sense it is trivial to convert a BN into a MN, but there is no guarantee that the associated undirected graph can represent the exact same conditional independences~\cite[chapter 19]{murphy2012mlapp}. Both kinds of graphs have different sets of distributions that they can represent perfectly (i.e.\ for which they are a perfect map).

Dynamic Bayesian networks (DBNs) are a type of Bayesian network that can represent stochastic processes. Unlike their name suggests, DBNs model stationary processes, the ``dynamic'' refers to their stochastic nature. Such processes produce sequences of observable and hidden variables, where the sequence index is often a discrete representation of time. Therefore DBNs are naturally represented by chainlike graphs that exhibit a certain regularity in the dependency relationships between their variables. This regularity allows for a compact graphical representation and it is amply exploited by the associated algorithms for inference and learning.

Hidden Markov models\footnote{The naming of HMMs is rather unfortunate in the sense that they are not Markov networks, but Bayesian networks.} (HMMs)~\cite{ghahramani2001ijprai} are closely related to DBNs. The boundary between the two is fuzzy, and is not always consistently defined. We employ the strict definition that an HMM is a DBN with only one (multivariate) hidden variable and one visible variable. If multiple hidden or visible variables arise in the process to be modelled, they are combined to form a single compound variable. This allows to simplify the associated algorithms. The drawback of this approach is that independence relationships within these compound variables are not exploited, whereas DBNs can use them to speed up inference or they can be imposed while learning. A number of specific model architectures branded as HMMs (such as factorial HMMs~\cite{jordan1996nips}, hierarchical HMMs~\cite{fine1998ml} and hidden semi-Markov models~\cite{yu2010ai})
have been proposed in the literature throughout the years, in order to address these issues. According to our strict definition, we would call them DBNs instead.

Finally, the difference between propositional logic and first-order logic~\cite{russell2009aiama} is that the former constructs its formulas using true/false statements only, whereas the latter can additionally use quantified variables in its formulas. Through its use of existential ($\exists$) or universal ($\forall$) quantifiers, more generally applicable statements can be made.

\section{Unique capabilities of Markov logic networks}
One way to look at MLNs, is to see them as an extension of first-order logic with weights. A set of logical statements then no longer complies with a given knowledge base in a binary fashion (yes or no), but an intermediate degree of compliance can be attained. This allows the knowledge base to contain contradictory statements, which will never be fulfilled at the same time.

Thinking about MLNs from a more probabilistic point of view, they can be seen as a way to describe Markov networks (MNs) using first-order logic. One could argue that standard MNs, or more generally probabilistic graphical models (PGMs) (comprising BNs and MNs), already are able to express relationships through the edges in their graphs and they handle uncertainty by associating probabilities to the edges or cliques. The difference is that edges in a PGM can be seen as propositional logic, a simple list of statements that describe which edges are present. If there is a recurring structure in a graph, it would be useful to include this recurrence into the graph description instead of giving an exhaustive list of edges. We could then, for instance, simply state that all nodes of a certain type need to be interconnected. Unfortunately, no mechanism to describe recurrence exists in ordinary PGMs. This is where the added expressiveness of the first-order logic (FOL) in MLNs comes in handy.

To understand the benefits of formulating a graph in FOL, it can be helpful to draw a parallel with the link between Bayesian networks (BNs) and dynamic Bayesian networks (DBNs). Any given DBN does not model its underlying stochastic process for one specific sequence length only, but for any sequence length. If we want to model the same process using an ordinary BN, we would need to create multiple networks, one for each of the process lengths under consideration. In other words, a DBN adapts seamlessly to the length of observed sequences that are presented to it. The advantage of DBNs over ordinary BNs is therefore twofold. First, it allows a compact and elegant formulation of a whole set of BNs that exhibit a specific regularity through time. Second, that regularity can be exploited by inference and learning algorithms that are more optimal than the algorithms used for general BNs. In contrast, the drawback of a DBN is that it can only describe a limited subset of the number of possible BNs: only those that are created through repeating a set of nodes as many times as necessary, a process known as \emph{unrolling} the network. The descriptive power of a DBN is therefore higher than propositional logic. It can be considered as a small subset of first-order logic, where only the sequence index variable can be universally quantified ($\forall$).

The advantages of MLNs over MNs are similar to the advantages of DBNs over BNs: both provide a convenient way to describe some regularity in a network, such that the formulation is suitable for observations of varying dimensions and such that the regularity can be exploited by the algorithms. MLNs are more broadly applicable than DBNs though: any regularity in an arbitrary number of dimensions can be described because the full descriptive power of first-order logic can be used. Any MN can therefore be described as an MLN: in case no regularity is present, the description will only use propositional logic. Obviously, the advantage of using MLNs is only noticeable when describing systems that cannot easily be described with propositional logic.

Multiple toolkits for MLNs are available that turn the underlying ideas into useable software. Among them is the reference implementation developed by the creators of the MLN theory, called Alchemy (version 2.0)\footnote{\url{https://alchemy.cs.washington.edu/}}. Its latest update stems from January 2013, however. A more recent toolkit is named ProbCog\footnote{\url{https://github.com/opcode81/ProbCog}} (last update July 2019), originating at the Technical University of Munich. Different toolkits implement different logic engines though, which means that their syntax and capabilities can slightly differ. Since the example MLNs given by Papadopoulos and Tzanetakis~\cite{papadopoulos2012ismir,papadopoulos2013icassp} are solved by the ProbCog toolkit, we will consider that variant in the remainder of this paper.

One remark is that research on MLN first focussed on completing the functionality for describing a MN through logic. The potential advantage of improving the inference and learning algorithms -- by exploiting the regularity present in the logic description -- is much more of a work-in-progress. A useful fallback option is to instantiate the underlying graphical model first in a process called grounding (comparable to the unrolling of a DBN into a BN by copy-pasting the repeated part for each sequence step) and then reuse the existing algorithms for standard Markov networks. Performing inference of MLNs without the need to ground the network first, is called \emph{lifted inference} and is the subject of ongoing research~\cite{braz2005ijcai,singla2008aaai,gogate2011uai}. 

Some of the first lifted inference algorithms are available in the Alchemy software, but the ProbCog toolkit only contains algorithms that work on grounded networks. Papadopoulos and Tzanetakis advocate the use of toulbar2~\footnote{\url{http://www.inra.fr/mia/T/toulbar2}}, an exact solver for Markov networks and other cost function networks included with ProbCog. The drawback is that one of the theoretical advantages of MLN over MN disappears in practice. This leaves us with the compact and elegant notation as main benefit of using MLNs. In the following sections, we will discuss why the particular requirements of music signal analysis make MLN formulations less elegant than they would be for other application domains.


\section{Disadvantages of MLNs as implemented in the ProbCog toolkit}

In a typical music signal analysis task, the goal is to take an audio signal as input and return one or more labels that describe the content of the music. In the example systems given by Papadopoulos and Tzanetakis, these labels are chords, a time-varying description of harmony. It is therefore natural to model them as a chainlike graph generated by a stationary process. The lack of explicit support for sequences in ProbCog is therefore unfortunate and forces us to use less elegant workarounds. Another typical characteristic of music analysis systems is that the observations are generally continuous (because they are in one way or another derived from a continuous waveform). Specifically for chord analysis, the observations are often a 12-dimensional time-frequency representation where frequencies are reduced to a single octave, known as a chromagram~\cite{wakefield1999spie,fujishima1999icmc}. As as result, the fact that ProbCog does not support continuous probability distributions is another drawback that requires a workaround.

\subsection{Handling continuous observation probability distributions}
The solution to handle continuous probability distributions proposed by Papadopoulos and Tzanetakis is to precalculate the probabilities of all observations in a specific song (as opposed to specifying the probabilities for all possible observations as a discrete observation distribution). The set of observations per song is then used as the discrete domain and each ``discrete value''  is encountered exactly once. The weighted formulas (notated as a numerical weight separated by whitespace from a logic formula) they propose for modelling the observations have the form
\begin{multline}
\operatorname{weight}\left(Label,N\right) \\ \operatorname{observation}\left(Obs_{N}, t\right) \wedge \operatorname{chord}\left(Label, t\right)\label{eq:observation-formula}
\end{multline} with $Obs_{N}$ the name given to the observations at index $N$.
We use the convention that logical variables are represented by lowercase text (such as $t$) and logical constants by capitalised text (such as $Label$). In practice, such a formula needs to be specified for every combination of the label and observation constants because the weights are dependent on both. The logical constants then take values such as \emph{Cmaj} and \emph{Amin} for $Label$ and $0,1,\ldots$ for $N$. Papadopoulos and Tzanetakis employ 24 chord labels and with a typical song being divided into hundreds of beats (time segments), the number of formulas that needs to be specified runs into the thousands.

The corresponding evidence needs to be provided for every song in the form of
\begin{equation}
\operatorname{observation}\left(Obs_N, N\right)\label{eq:observation-evidence}
\end{equation}
which effectively says that ``the observation at time $N$ is called the observation at time $N$''. The whole process entails the creation of a custom MLN per audio file under analysis, instead of defining one model that can adapt to any sequence of observations, as would be the case for true discrete observations.

Because each of the discretised observations is guaranteed to appear only once in the sequence, the validity of the FOL observation formulas is limited to singleton domains. It is superfluous to specifically create named constants $Obs_N$ just to link the time sequence index $t$ to the index of the discrete set of observations $n$ (they are always equal). The combination of formulas of the form (\ref{eq:observation-formula}) and evidence of the form (\ref{eq:observation-evidence}) can be simplified by substituting the sequence index variable $t$ by the only valid index $N$ in the $\operatorname{observation}$ predicate. The resulting formulas are then of the form
\begin{equation}
\operatorname{weight}\left(Label,N\right) \qquad \operatorname{observation}\left(N\right) \wedge \operatorname{chord}\left(Label, t\right)
\end{equation}
with evidence of the form
\begin{equation}
\operatorname{observation}\left(N\right)
\end{equation}

Now the sole reason of existence for the $\operatorname{observation}$ predicate is to pass on the index $N$ to the variable $t$ in the $\operatorname{chord}$ predicate, so we can remove the former altogether by substituting $t$ by $N$ again. The final formula is then
\begin{equation}
\operatorname{weight}\left(Label,N\right) \qquad \operatorname{chord}\left(Label, N\right)
\end{equation}
with no need to specify evidence. This formulation is not only much simpler than the originally proposed one, it also is in propositional logic form, showing that the added complexity of the first-order logic formulation is unnecessary in this case, and only obfuscates the formulation.

In summary, using continuous observation variables demands that the observation probabilities are calculated externally and that they are written in time-unrolled (propositional) form into the MLN configuration on a file-per-file basis. Consequently, the separation between logical network definition and evidence is compromised and most of the elegance of the logical description gets lost. The actual file-dependent observations get absorbed into the model definition instead of being evidence. Looking beyond the ProbCog toolkit, some work has been done to extend the Alchemy MLN toolkit with continuous observation distributions~\cite{wang2008aaai}, but those distributions are limited to Gaussians. Since the observation probabilities in~\cite{papadopoulos2012ismir,papadopoulos2013icassp,papadopoulos2017taslp} are calculated as the correlation between chroma vectors and theoretical templates, using Alchemy is not a viable solution.


\subsection{Handling sequences}

The underlying problem that makes sequences hard to handle in ProbCog, is that there is no support for arithmetic operators between variables. This absence becomes especially notable when we want to encode the probabilities of pairwise transitioning between chords (such as in an HMM) into formulas. We want to specify that the probabilities only apply to two timestamps $t_1$ and $t_2$ if they are related as $t_2=t_1+1$. The way this problem is solved in~\cite{papadopoulos2012ismir,papadopoulos2013icassp,papadopoulos2017taslp}, is by adding an extra helper predicate (here called $\operatorname{next}/2$) to the logic formula to control when the formula is valid:
\begin{multline}\label{eq:transition-formula}
\operatorname{weight}\left(Label_1,Label_2\right) \\ \operatorname{chord}\left(Label_1, t_1\right) \wedge \operatorname{chord}\left(Label_2, t_2\right) \wedge \operatorname{next}\left(t_1, t_2\right)
\end{multline}
It is then necessary to provide evidence of the form $\operatorname{next}(N,N+1)$ as many times as necessary for the song length. So just like for the observation formulas, this requires manual unrolling of time.

Looking at it graphically, this means that the (hidden) chord nodes of all sequence indices in the model are fully interconnected. In order to ``activate'' the connection however, the associated predicate $\operatorname{next}\left(N_1,N_2\right)$ between a specific pair of timestamps $\left(N_1,N_2\right)$ needs to be given as evidence\footnote{We assume a closed world, such that withholding evidence is the same as giving evidence of its negation}. Only in that case does the size three clique formed by formula \eqref{eq:transition-formula} become true. 
The model definition itself thus leaves the possibility of non-linear time open, e.g.\ 4--4--2--6--5--1--4, and it relies on the given evidence to make it linear, e.g.\ 1--2--3--4--5--6. Again, we see that the separation between model definition and evidence gets blurred, because the model relies on specific evidence to be given in order to complete the network definition.

Furthermore, the $\operatorname{next}/2$ predicate cannot be reused to make additional, arbitrary connections between non-consecutive chord nodes (for instance to probabilistically tie repetitions of the same chord subsequence together, as proposed in~\cite{papadopoulos2013icassp}). Specifying $\operatorname{next}\left(t_1,t_3\right),\forall t_3\neq t_1+1$ as evidence does create such an additional connection, but its associated weight would still encode the pairwise probability between consecutive chords, leading to unintended and incorrect transition probabilities. Any additional connection then requires the introduction of another $\operatorname{next}/2$-like predicate, connecting all chord node combinations together once more in supplementary cliques of 3. The number of nodes and connections in the network rapidly soars this way.

In short, the absence of arithmetic operators in ProbCog makes a FOL network description instantiate suboptimal graphs. The result is fully connected and relies on the presence of evidence to prune the nonsensical connections. This approach is way more convoluted than defining only those connections that need to be made in the first place. An alternative to accomplish the latter, would be to get rid of the $\operatorname{next}/2$ predicate and write formulas down in propositional logic form only when a connection between the two sequence indices $T_1$ and $T_2$ is actually required. The form will then be
\begin{multline}
\operatorname{weight}\left(Label_1,Label_2\right) \\ \operatorname{chord}\left(Label_1, T_1\right) \wedge \operatorname{chord}\left(Label_2, T_2\right)
\end{multline}
without any corresponding evidence. The drawback of specifying the transition probabilities through propositional logic, is that the number of formulas for the model configuration will increase. But because the configuration needs to be song-dependent and time-unrolled for previously mentioned reasons, the configuration file realistically needs to be generated programmatically anyway. Therefore, this is not an issue in practice.


Alchemy does have partial support for arithmetic operators, but our desired use-case -- in which the transition probabilities would be elegantly formulated in the FOL form below -- is not supported.
\begin{multline}
\operatorname{weight}\left(Label_1,Label_2\right) \\ \operatorname{chord}\left(Label_1, t\right) \wedge \operatorname{chord}\left(Label_2, t+1\right)
\end{multline}
In addition, Alchemy does not include an exact inference algorithm.

\section{Alternative formulations of concrete MLN examples}

The difference between the MLN systems proposed in~\cite{papadopoulos2012ismir,papadopoulos2013icassp,papadopoulos2017taslp} and other work on automatic chord estimation is twofold: the former can use the full expressive power of first-order logic to describe relations instead of the more limited expressiveness of the latter, and the underlying graphical model is a Markov network instead of a Bayesian network. We have shown in the previous section that in practice, the full power of FOL cannot be used when describing chord estimation systems. In this section, we will reformulate the concrete MLN examples as (dynamic) Bayesian networks to demonstrate that their underlying ideas can be expressed in this more conventional framework as well. It also makes it easier to compare the proposed approaches to other systems, for example to previous work integrating key and chord estimation~\cite{mauch2010taslp,pauwels2014jnmr}.

Common to all systems in~\cite{papadopoulos2012ismir,papadopoulos2013icassp,papadopoulos2017taslp} is that they focus on inference. The weights are set manually instead of being learned. We can divide the examples into two groups. The first contains the three systems first described in~\cite{papadopoulos2012ismir} and later repeated in~\cite{papadopoulos2017taslp}. The first system estimates chords based on chroma observations, the second additionally takes prior key information into account and the third performs joint key and chord estimation. All three systems have weights that are derived from conditional probabilities and only consider pairwise time dependencies. Added to the fact that all formulas in the MLNs are conjunctions of positive literals, this means that the resulting MNs instantiated from the MLNs can be equally well represented as DBNs~\cite{sang2005aaai}.


\begin{figure}[htb]
\medmuskip=0mu
\begin{minipage}[t]{0.4\linewidth}
  \centering
  \begin{tikzpicture}[every node/.append style={align=center, text width=1cm}]
	  \node[latent] (c1) {$\operatorname{c}\left(t-1\right)$};
	  \node[latent, right=6mm of c1] (c2) {$\operatorname{c}\left(t\right)$};
	  \node[obs, below=6mm of c2] (o) {$\operatorname{o}\left(t\right)$};
	
	  \edge {c1} {c2} ; %
	  \edge {c2} {o} ; %
	  
  \end{tikzpicture}
  \centerline{(a) chord estimation}\medskip
\end{minipage}
\begin{minipage}[t]{0.6\linewidth}
  \centering
  \begin{tikzpicture}[every node/.append style={align=center, text width=1cm}]
	  \node[latent] (c1) {$\operatorname{c}\left(t-1\right)$};
	  \node[latent, right=6mm of c1] (c2) {$\operatorname{c}\left(t\right)$};
	  \node[obs, below=6mm of c2, xshift=-7mm] (o) {$\operatorname{o}\left(t\right)$};
	  \node[obs, below=6mm of c2, xshift=7mm] (k) {$\operatorname{k}\left(t\right)$};
	
	  \edge {c1} {c2} ; %
	  \edge {c2} {o,k} ; %
	  
  \end{tikzpicture}
  \centerline{(b) chord estimation with prior key}\medskip
\end{minipage}

\begin{minipage}[t]{\linewidth}
  \centering
  \begin{tikzpicture}[every node/.append style={align=center, text width=1cm}]
	  \node[latent] (k1) {$\operatorname{k}\left(t-1\right)$};
	  \node[latent, below=6mm of k1] (c1) {$\operatorname{c}\left(t-1\right)$};
	  \node[latent, right=6mm of k1] (k2) {$\operatorname{k}\left(t\right)$};
	  \node[latent, right=6mm of c1] (c2) {$\operatorname{c}\left(t\right)$};
	  \node[obs, below=6mm of c2] (o) {$\operatorname{o}\left(t\right)$};
	
	  \edge {k1} {k2} ; %
	  \edge {k1} {c1} ; %
	  \edge {k2} {c2} ; %
	  \edge {c1} {c2} ; %
	  \edge {c2} {o} ; %
	  
  \end{tikzpicture}
  \centerline{(c) joint key and chord estimation}\medskip
\end{minipage}
\caption{Equivalent DBN representations of the systems presented in~\cite{papadopoulos2012ismir}, where $\operatorname{o}\left(t\right)$, $\operatorname{c}\left(t\right)$ and $\operatorname{k}\left(t\right)$ stand for respectively the chroma observation, chord and key at time $t$. Shaded nodes represent observed variables and white nodes are hidden variables.}
\label{fig:dbns}
\end{figure}

\begin{figure}[htb!]
\begin{minipage}[b]{\linewidth}
  \centering
  \begin{tikzpicture}
	  \node[latent] (c0) {$\operatorname{c}\left(0\right)$};
	  \node[latent,right=6mm of c0] (c1) {$\operatorname{c}\left(1\right)$};
	  \node[latent,right=6mm of c1] (c2) {$\operatorname{c}\left(2\right)$};
	  \node[latent,right=6mm of c2] (c3) {$\operatorname{c}\left(3\right)$};
	  \node[latent,right=6mm of c3] (c4) {$\operatorname{c}\left(4\right)$};
	  \node[latent,right=6mm of c4] (c5) {$\operatorname{c}\left(5\right)$};
	  \node[obs,below=6mm of c0] (o0) {$\operatorname{o}\left(0\right)$};
	  \node[obs,below=6mm of c1] (o1) {$\operatorname{o}\left(1\right)$};
	  \node[obs,below=6mm of c2] (o2) {$\operatorname{o}\left(2\right)$};
	  \node[obs,below=6mm of c3] (o3) {$\operatorname{o}\left(3\right)$};
	  \node[obs,below=6mm of c4] (o4) {$\operatorname{o}\left(4\right)$};
	  \node[obs,below=6mm of c5] (o5) {$\operatorname{o}\left(5\right)$};
	
	  \edge {c0} {c1} ;
	  \edge {c1} {c2} ;
	  \edge {c2} {c3} ;
	  \edge {c3} {c4} ;
	  \edge {c4} {c5} ;
	  \edge {c0} {o0} ;
	  \edge {c1} {o1} ;
	  \edge {c2} {o2} ;
	  \edge {c3} {o3} ;
	  \edge {c4} {o4} ;
	  \edge {c5} {o5} ;
	  \draw[->] (c0) [bend left] to  (c3);
	  \draw[->] (c1) [bend left] to  (c4);
	  \draw[->] (c2) [bend left] to (c5);
	  
  \end{tikzpicture}
  \centerline{(a) BN equivalent for ``MLNStruct''}\medskip
\end{minipage}

\begin{minipage}[b]{\linewidth}
  \centering
  \begin{tikzpicture}
	  \node[latent] (c0) {$\operatorname{c}\left(0\right)$};
	  \node[latent,right=6mm of c0] (c1) {$\operatorname{c}\left(1\right)$};
	  \node[latent,right=6mm of c1] (c2) {$\operatorname{c}\left(2\right)$};
	  \node[latent,right=6mm of c2] (c3) {$\operatorname{c}\left(3\right)$};
	  \node[latent,right=6mm of c3] (c4) {$\operatorname{c}\left(4\right)$};
	  \node[latent,right=6mm of c4] (c5) {$\operatorname{c}\left(5\right)$};
	  \node[obs,below=6mm of c0] (o0) {$\operatorname{o}\left(0\right)$};
	  \node[obs,below=6mm of c1] (o1) {$\operatorname{o}\left(1\right)$};
	  \node[obs,below=6mm of c2] (o2) {$\operatorname{o}\left(2\right)$};
	  \node[obs,below=6mm of c3] (o3) {$\operatorname{o}\left(3\right)$};
	  \node[obs,below=6mm of c4] (o4) {$\operatorname{o}\left(4\right)$};
	  \node[obs,below=6mm of c5] (o5) {$\operatorname{o}\left(5\right)$};
	  \node[obs,above=6mm of c0] (k0) {$\operatorname{k}\left(0\right)$};
	  \node[obs,above=6mm of c1] (k1) {$\operatorname{k}\left(1\right)$};
	  \node[obs,above=6mm of c2] (k2) {$\operatorname{k}\left(2\right)$};
	  \node[obs,above=6mm of c3] (k3) {$\operatorname{k}\left(3\right)$};
	  \node[obs,above=6mm of c4] (k4) {$\operatorname{k}\left(4\right)$};
	  \node[obs,above=6mm of c5] (k5) {$\operatorname{k}\left(5\right)$};
	
	  \edge {c0} {c1} ;
	  \edge {c1} {c2} ;
	  \edge {c2} {c3} ;
	  \edge {c3} {c4} ;
	  \edge {c4} {c5} ;
	  \edge {c0} {o0} ;
	  \edge {c1} {o1} ;
	  \edge {c2} {o2} ;
	  \edge {c3} {o3} ;
	  \edge {c4} {o4} ;
	  \edge {c5} {o5} ;
	  \draw[->] (c0) [bend left] to  (c3);
	  \draw[->] (c1) [bend left] to  (c4);
	  \draw[->] (c2) [bend left] to (c5);
	  \edge {c0} {k0} ;
	  \edge {c1} {k1} ;
	  \edge {c2} {k2} ;
	  \edge {c3} {k3} ;
	  \edge {c4} {k4} ;
	  \edge {c5} {k5} ;
	  
  \end{tikzpicture}
  \centerline{(b) BN equivalent for ``MLNMultiScale-PriorKey''}\medskip
\end{minipage}

\begin{minipage}[b]{\linewidth}
  \centering
  \begin{tikzpicture}[align=center]
	  \node[latent] (c0) {$\operatorname{c}\left(0\right)$};
	  \node[latent,right=6mm of c0] (c1) {$\operatorname{c}\left(1\right)$};
	  \node[latent,right=6mm of c1] (c2) {$\operatorname{c}\left(2\right)$};
	  \node[latent,right=6mm of c2] (c3) {$\operatorname{c}\left(3\right)$};
	  \node[latent,right=6mm of c3] (c4) {$\operatorname{c}\left(4\right)$};
	  \node[latent,right=6mm of c4] (c5) {$\operatorname{c}\left(5\right)$};
	  \node[obs,below=6mm of c0] (o0) {$\operatorname{o}\left(0\right)$};
	  \node[obs,below=6mm of c1] (o1) {$\operatorname{o}\left(1\right)$};
	  \node[obs,below=6mm of c2] (o2) {$\operatorname{o}\left(2\right)$};
	  \node[obs,below=6mm of c3] (o3) {$\operatorname{o}\left(3\right)$};
	  \node[obs,below=6mm of c4] (o4) {$\operatorname{o}\left(4\right)$};
	  \node[obs,below=6mm of c5] (o5) {$\operatorname{o}\left(5\right)$};
	  \node[latent,above=6mm of c0] (k0) {$\operatorname{k}\left(0\right)$};
	  \node[latent,above=6mm of c1] (k1) {$\operatorname{k}\left(1\right)$};
	  \node[latent,above=6mm of c2] (k2) {$\operatorname{k}\left(2\right)$};
	  \node[latent,above=6mm of c3] (k3) {$\operatorname{k}\left(3\right)$};
	  \node[latent,above=6mm of c4] (k4) {$\operatorname{k}\left(4\right)$};
	  \node[latent,above=6mm of c5] (k5) {$\operatorname{k}\left(5\right)$};
	
	  \edge {c0} {c1} ;
	  \edge {c1} {c2} ;
	  \edge {c2} {c3} ;
	  \edge {c3} {c4} ;
	  \edge {c4} {c5} ;
	  \edge {c0} {o0} ;
	  \edge {c1} {o1} ;
	  \edge {c2} {o2} ;
	  \edge {c3} {o3} ;
	  \edge {c4} {o4} ;
	  \edge {c5} {o5} ;
	  \draw[->] (c0) [bend left] to (c3);
	  \draw[->] (c1) [bend left] to (c4);
	  \draw[->] (c2) [bend left] to (c5);
	  \draw[->] (k0) [bend left] to (k1);
	  \draw[->] (k1) [bend left] to (k2);
	  \draw[->] (k3) [bend left] to (k4);
	  \draw[->] (k4) [bend left] to (k5);
	  \edge {k0} {c0} ;
	  \edge {k1} {c1} ;
	  \edge {k2} {c2} ;
	  \edge {k3} {c3} ;
	  \edge {k4} {c4} ;
	  \edge {k5} {c5} ;
	  \edge {k0} {k1} ;
	  \edge {k1} {k2} ;
	  \edge {k2} {k3} ;
	  \edge {k3} {k4} ;
	  \edge {k4} {k5} ;
	  
  \end{tikzpicture}
  \centerline{(c) BN equivalent for ``MLNMultiScale''}\medskip
\end{minipage}

\caption{Equivalent BN representations of the more complex systems presented in~\cite{papadopoulos2017taslp}, where $\operatorname{o}\left(t\right)$, $\operatorname{c}\left(t\right)$ and $\operatorname{k}\left(t\right)$ stand for respectively the chroma observation, chord and key at time $t$. Shaded nodes represent observed variables and white nodes are hidden variables. It is assumed that nodes $[0:2]$ and $[3:5]$ belong to the same measure and that the time segment pairs $\left(0,3\right)$,$\left(1,4\right)$ and $\left(2,5\right)$ are given as highly similar.}
\label{fig:bn}
\end{figure}

The diagrams for those three systems are displayed in \figref{dbns}. The formulations as DBNs will actually be more compact and elegant because continuous observations densities can be used and the observation probabilities no longer need to be unrolled in time. Moreover, the DBN can be reused multiple times by changing the observations presented to it, whereas the MLNs need to be redefined for every song because of the conflation between model and evidence. Another advantage of this formulation is that optimised inference algorithms can be used. For instance, we can now see that it is not surprising that Papadopoulos and Tzanetakis found the difference between the chord estimation MLN and a chord HMM to be statistically not significant~\cite{papadopoulos2012ismir}, as they describe exactly the same system. The only difference is that the MLN description needs to resort to a more generally applicable inference algorithm because of the time-unrolled observations and suboptimal network layout, whereas the HMM can use the Viterbi algorithm for decoding (resulting in a speedup of orders of magnitude).


The second group of MLN examples is arguably more interesting, precisely because they cannot be formulated as DBNs (but still as BNs). It comprises three of the systems presented in~\cite{papadopoulos2017taslp}, one of which was previously presented in~\cite{papadopoulos2013icassp}. The motivation behind them is the fact that one drawback of DBNs is their inability to take longer term dependencies between chords directly into account. We assume that we dispose of a number of pairs that indicate high similarity between its (non-adjacent) constituents. These similarity pairs can come from a prior structural analysis under the assumption that a repetition of a chorus, for instance, will contain the same chords as the previous chorus. Papadopoulos and Tzanetakis use manual structural annotations to derive these similarity pairs, such that they are binary (either similar or not). The repetitions are then exploited to improve the chord estimation. In this sense, this approach is comparable to other systems that take into account repetition, such as~\cite{mauch2009ismir,cho2011ismir}. The difference is that here the knowledge about the repetitions is integrated probabilistically into the inference, whereas those previous approaches do a deterministic early fusion of the repeated observations.



In essence, the three systems are constructed by unrolling the previous three DBN systems through time such that the similarity knowledge can be integrated by drawing additional edges between the similar time segments. Example graphs with six time segments are shown in \figref{bn}. The most complex example not only takes long term chord similarity based on structure into account, but also medium term key similarity by connecting keys in the same measure under the assumption that key changes within a measure are highly unlikely.

Inference in the resulting BNs will not be faster than in the optimal, propositionally described MNs (which have the same graphs, but with undirected edges instead), but the BNs have the advantage of being easier to interpret. The weights proposed in~\cite{papadopoulos2017taslp} for the similarity-based edges are fixed without justification or interpretation, presumably found through a parameter sweep. Because of the global normalisation in MN, this makes it hard to interpret the numbers. Even worse, the interpretation of the other weights (derived from conditional probabilities) will be affected by the addition of the supplemental structural edges, which are by definition unpredictable and song-dependent, and depend on the given evidence~\cite{jain2011ki}. The whole graph then quickly becomes opaque. In a BN, the structural edges will get a conditional probability assigned, which has to be normalised locally to preserve the total probability in each node. The optimal parameter still can -- and likely has to -- be found through a parameter sweep, but the result will be interpretable and will not affect the other parameters in unforeseeable ways. A BN will also make it easier, for example, to consolidate the key transition probabilities and the probabilistic tying of keys within the same measure in the last graph of \figref{bn} into a single edge (currently they are represented by separate edges connecting the same nodes).


%
%


\section{Conclusion and future work}
In this paper, we studied MLNs for music analysis by analysing a couple of previously proposed example systems for chord estimation and reformulating them as (dynamic) Bayesian networks. The theoretical advantages of MLN are an easier, more powerful formulation of graphical models and an improvement in learning/inference algorithms through exploiting regularity in the networks. In practice, however, we saw that the MLN systems formulated in~\cite{papadopoulos2012ismir,papadopoulos2013icassp,papadopoulos2017taslp} are not optimally exploiting the capabilities of the MLN framework. This is partly because these formulations introduce unnecessary variables to create FOL where propositional logic would suffice, and partly because the lack of support for continuous observation distributions and sequences forces the use of convoluted workarounds. We furthermore showed that it is not necessary to resort to Markov networks to express the underlying ideas, but that Bayesian networks can use the same principles, with improved interpretability. We argue that because music signals are variable through time and causal, a directional left-to-right relationship arises naturally when modelling music,therefore the associated graphs will be chainlike. These conclusions are not specific to music, but apply to other fields as well, especially when sequential data with continuous observations is involved.

This is not to say that MLNs are not useful for solving a number of problems, even in the musical domain. Candidate problems for which MLNs would provide a good fit have only discrete variables and would ideally model static processes. Any data that is topographically organised would be suitable, as such a layout increases the chance of cyclical relationships and regularity that cannot be properly modelled by (dynamic) Bayesian networks. In the musical domain, one application that fulfils these requirements would be similarity discovery or clustering between artists, as long as no time-varying features are used.

Another case where MLNs can be helpful is when a large body of knowledge about the problem at hand already exists in first-order logic form. In contrast, the formulas in~\cite{papadopoulos2012ismir,papadopoulos2013icassp,papadopoulos2017taslp} are relatively simple, just describing relationships between node pairs. In this scenario, the related paradigm of Bayesian logic networks (BLNs)~\cite{jain2009tr} might be an alternative if no bidirectional connections are needed. An implementation is available in the same ProbCog toolkit, which has the advantage that arithmetic are supported such that it is possible to impose $t_2=t_1+1$ through a logical rule. Continuous observation distributions require the same workaround as with MLNs though.

Another alternative for MLNs we could explore in the future is ProbLog~\cite{fierens2015tplp}. The advantage of this approach is that it is built around a complete Prolog engine, which makes it easy to extend with custom functionality, such as arbitrary continuous distributions. In our first preliminary experiments, however, we already encountered problems to make it scale to the typical network sizes of our problems, despite recent advances in this aspect~\cite{vlasselaer2016ai}.

\section*{Acknowledgements}
This work has been partly funded by the UK Engineering and Physical Sciences Research Council (EPSRC) grant EP/L019981/1 and by the European Union’s Horizon 2020 research and innovation programme under grant agreement N$^\circ$ 688382.

\bibliographystyle{aaai}
\bibliography{strings,pauwels2020starai}

\end{document}